%% file: acl_latex.tex
\pdfoutput=1
\documentclass[11pt]{article}

\usepackage[preprint]{acl}
\usepackage{amsmath}
\usepackage{times}
\usepackage{booktabs}
\usepackage{latexsym}
\usepackage{enumitem}

\usepackage[T1]{fontenc}
\usepackage[utf8]{inputenc}
\usepackage{CJKutf8}
\usepackage{microtype}
\usepackage{multirow}
\usepackage{tabularx}
\usepackage{makecell}

\usepackage{inconsolata}

\usepackage{graphicx}

\title{Context Graph}

\author{Chengjin Xu$^{1}$\footnotemark[1], Muzhi Li$^{1,2}$\footnotemark[1] , Cehao Yang$^{1}$, Xuhui Jiang$^{1,3}$, Lumingyuan Tang$^{1}$, Yiyan Qi$^{1}$, \\  \textbf{Jian Guo}$^{1}$\footnotemark[2] \\
1. IDEA Research, International Digital Economy Academy \\
  2. 	Department of Computer Science and Engineering, The Chinese University of Hong Kong \\
  3. CAS Key Laboratory of AI Safety, Institute of Computing Technology, CAS \\
  \texttt{\{xuchengjin,limuzhi,yangcehao,jiangxuhui, guojian\}@idea.edu.cn} 
  }

\begin{document}
\maketitle
\renewcommand{\thefootnote}{\fnsymbol{footnote}}
\footnotetext[1]{Both authors contributed equally to this research.}
\footnotetext[2]{Corresponding author.}
\begin{abstract}
Knowledge Graphs (KGs) are foundational structures in many AI applications, representing entities and their interrelations through triples. However,  triple-based KGs lack the contextual information of relational knowledge, like temporal dynamics and provenance details, which are crucial for comprehensive knowledge representation and effective reasoning. Instead, \textbf{Context Graphs} (CGs) expand upon the conventional structure by incorporating additional information such as time validity, geographic location, and source provenance. This integration provides a more nuanced and accurate understanding of knowledge, enabling KGs to offer richer insights and support more sophisticated reasoning processes. In this work, we first discuss the inherent limitations of triple-based KGs and introduce the concept of CGs, highlighting their advantages in knowledge representation and reasoning. We then present a context graph reasoning \textbf{CGR$^3$} paradigm that leverages large language models (LLMs) to retrieve candidate entities and related contexts, rank them based on the retrieved information, and reason whether sufficient information has been obtained to answer a query. Our experimental results demonstrate that CGR$^3$ significantly improves performance on KG completion (KGC) and KG question answering (KGQA) tasks, validating the effectiveness of incorporating contextual information on KG representation and reasoning.

\end{abstract}

\input{intro.tex}

\input{definition}
\begin{figure*}
    \centering
    \includegraphics[width=.9\textwidth]{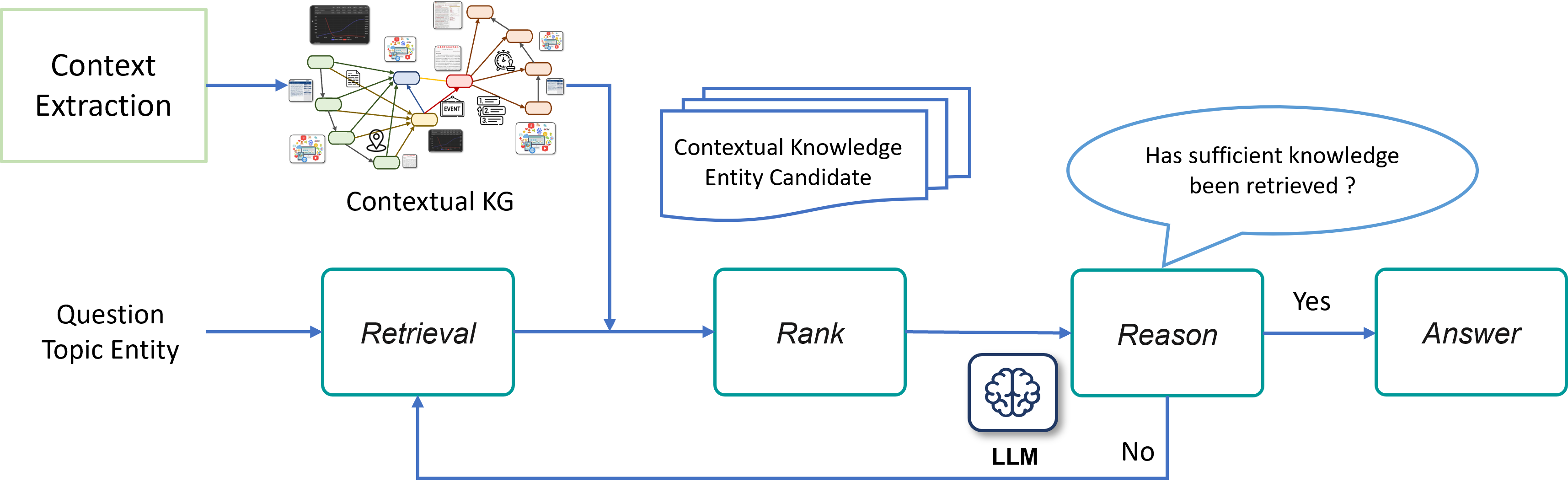}
    \caption{The pipeline of the CGR$^3$ paradigm. }
    \label{KGRpipe}
\end{figure*}

\section{Methods}
% 这个部分要改，我们没有空间像ChatEA一样写。
In this section, we introduce CGR$^3$, a novel context graph reasoning paradigm that leverages LLMs to perform knowledge reasoning tasks based on structured and contextual semantics. 
We aim to utilize the complementary relationship between both semantics to improve the reliability and explainability of the reasoning process.

As shown in For triple-based KGs, we begin by augmenting the KG with necessary contextual information extracted from relevant databases, a step that can be omitted if the KG is already a CG. The CGR$^3$ paradigm consists of three main steps: The \textbf{Retrieval} step is to retrieve candidate entities and related contexts from the CG based on the given question; the \textbf{Ranking} step involves ranking candidate entities based on the contexts and the given question; the \textbf{Reasoning} step is to exploit the LLM to determine whether sufficient information is retrieved. If sufficient information is available, the answer will be generated. If not, the whole processes iterates by retrieving new information based on the top-ranked candidate entities. We give a detailed description of the proposed context-aware paradigms for the KGC and the KBQA tasks
% 这部分不要了，但里面的句子可以用在别的地方
% \begin{itemize}
%     \item \textbf{O1: Augmenting KGs with necessary contextual semantics: } This objective aims to equip downstream models with a fundamental understanding of entities and triples in KGs.
%     %, particularly for boundary entities that have few neighbors. 
%     %(a low degree).
%     %告诉KGC模型或者KGR模型，这个entity是什么，尤其是那些边缘的（知识图谱是long-tailed的）
%     \item \textbf{O2: Enriching KGs with more plausible structural triples: } We aim to explore an effective KGC paradigm that overcomes the limitations of existing KGC methods, which predominantly rely on embedding similarity comparisons or distance measurements. 
%     % 在embedding based方法之外，探索一种新的KGC形式
%     \item \textbf{O3: Integrating contextual and structural semantics for enhanced KGR: } We aim to utilize the complementary relationship between the two types of semantics to improve the reliability and explainability of the reasoning process.
% \end{itemize}

\begin{figure*}
    \centering
    \includegraphics[width=\textwidth]{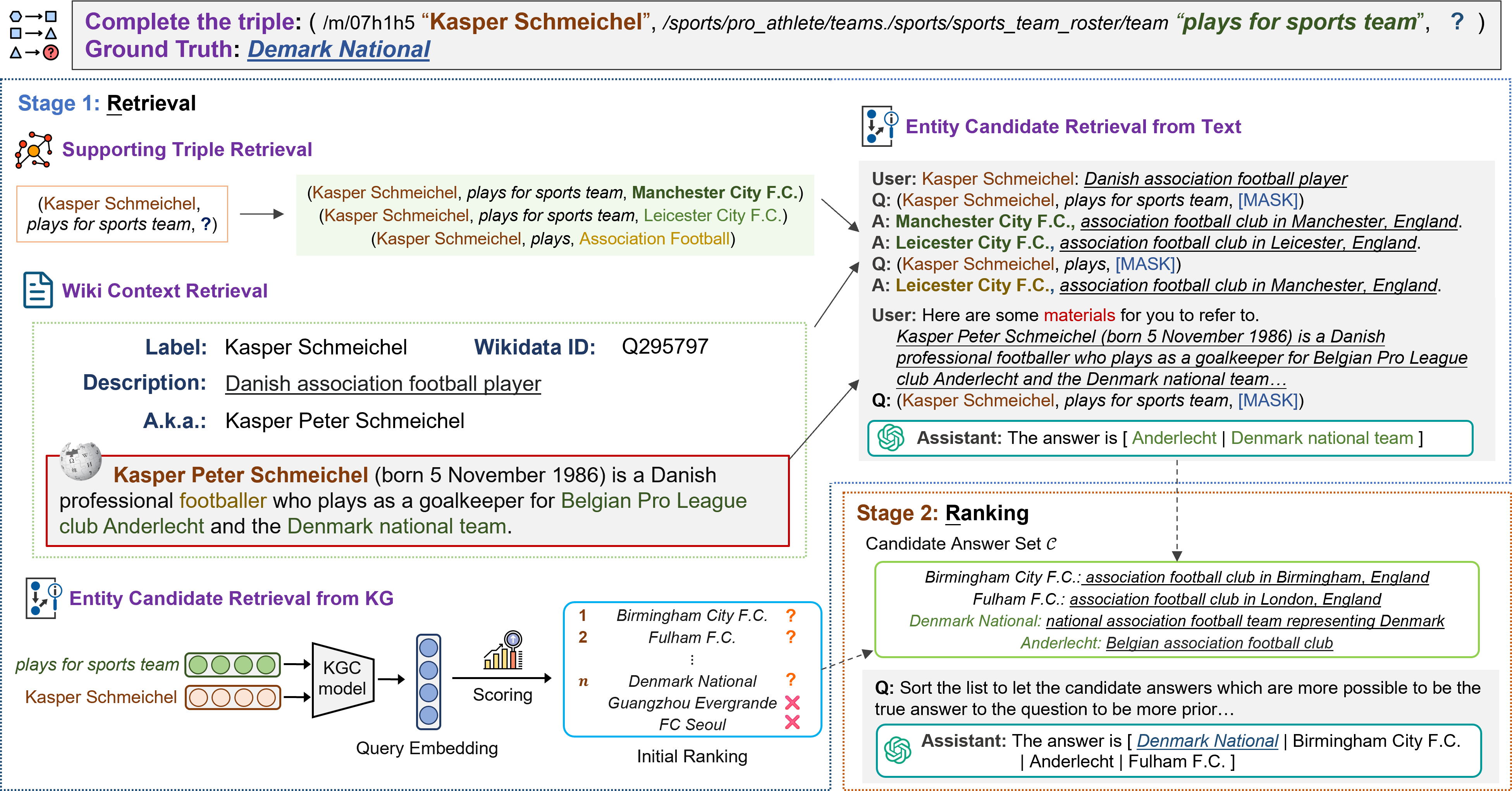}
    \caption{Knowledge Graph Completion. }
    \label{KGCFigure}
\end{figure*}

\subsection{Context Extraction}
\label{ContextExtraction}
% 为什么要抽取context：构建KG时候忽略了context
% 当前用context只是用label和description，这就不够。
% 为什么不够？因为它无法补充被忽略的triple-specific context
% 所以解决不了具体的KGR问题。

% 当前，普遍使用的知识图谱数据集，例如"FB15k237,YAGO3-10,Wikidata5M"是百科类图谱，它们反应了真实世界中的一般知识。 这些KG通常由领域领域专家在例如维基百科的知识库中通过识别实体和提取关系的方式构建。然而，在构建的过程中，有关实体的上下文信息经常被忽略。近期，一些研究工作讲... 然而，labels 和descriptions只能反映entities的一般知识，并不足以代替和具体triple相关的上下文信息。因此，它们并不足以用于handle diverse KGR problems
Currently, commonly used KG datasets, such as FB15k237, YAGO3-10, and Wikidata5M, are encyclopedic KGs that encapsulate general knowledge about the real world. These KGs are typically developed by domain experts by applying named entity recognition and relation extraction techniques on Wikipedia. However, during this construction process, the rich contexts surrounding the entities are often omitted.  
% 重新考虑这一句
Recent studies~\cite{KEPLER,CLMKE} have proposed to incorporate entity labels and descriptions as supplementary information for KGs. Nevertheless, the labels and descriptions are insufficient to replace the specific contexts associated with KG triples, thereby limiting their effectiveness in addressing diverse knowledge reasoning problems.

% 必要时，底下两段合并成一段。
% 我读起来总觉得怪怪的像有些东西讲了两遍
% 我换下脑子先写后面，之后再回来读这里。
To incorporate related contexts into KGs, we consider using Wikidata and Wikipedia as our primary contextual corpus in this work. Due to the extensive coverage and up-to-date information of Wikidata, some KGs like Freebase and YAGO provides official mapping files which can map their entities to Wikidata QIDs. For entities in other KGs, we can use entity search engines provided by Wikidata to find the Wikidata entities which are most likely to be identical to searched entities. Furthermore, Wikidata provides links to the associated Wikipedia pages of its entities. Thus, we can provide contextual information from Wikidata and Wikipedia to different KGs.

\subsubsection{Entity Context Extraction}
We start to complement the context of a KG with its entities. 
Specifically, we map the entities from Freebase, YAGO or other KBs to Wikidata QIDs by using official mapping files or using entity search engine provided by Wikidata. For each entity $e_{i}\in\mathcal{E}$, we collect the textual entity label, the short description, and aliases from Wikidata URIs as its entity context $ec_{i}\in\mathcal{EC}$. Moreover, the associated Wikipedia pages of Wikidata entities offer vital contextual support for the entities in the KGs. For each entity $e_{i}$, we integrate the Wikipedia pages as a part of entity contexts $ec_i$. 

\subsubsection{Relation Context Extraction}
For each triple, we aggregate the Wikipedia pages of its head and tail entities into a single document. Subsequently, we utilize Sentence-BERT~\cite{sentence-bert} to identify top-$\gamma$ supporting sentences that best reflect the semantics of the triple from this document. These sentences not only restore the contexts omitted during the KG construction but also provide optimal support for language models in understanding the structured KG triples. Thus, we can regard these supporting sentences as a kind of provenance information or supplementary information and treat them as relation contexts of triples. On the other word, for each triple $(h,r,t)$, we use its supporting sentences extracted from Wikipedia as its relation context $rc\in \mathcal{RC}$ and reshape this triple to a context-aware quadruple $(h,r,t,rc)$.

\subsection{Knowledge Graph Completion}
% 别删， 这部分要放到related works里去。
% Existing KGC methods~\cite{TransE,RotatE,CompoundE} mainly focus on embedding entities and relations in the KG with pre-defined message-passing and scoring functions. However, due to information scarcity, KGs exhibit a pronounced long-tail characteristic~\cite{longtail,CLMKE,KICGPT}, where many entities appear infrequently. As a result, KG embeddings often fail to accurately represent all the dominant attributes of entities, especially those with few neighbors. This greatly affects the effectiveness of the widely used KGC paradigm. It is worth noting that several works~\cite{kgbert,KEPLER,SimKGC} attempt to encode entity embeddings with entity names and descriptions with language models. Nevertheless, despite undergoing expensive fine-tuning processes, these text-based KGC approaches still cannot outperform traditional embedding methods~\cite{CompoundE}. Inspired by the reasoning capability of LLMs~\cite{KICGPT}, we demonstrate a new KGC paradigm that integrates and utilizes contextual semantics from three different perspectives. 
In this section, we demonstrate a new context-enriched KGC method based on our proposed CGR$^3$ paradigm. Since KGC can be considered as an entity ranking task for single-hop reasoning questions, it is not necessary to perform iterative reasoning processes. Thus, the \textbf{reasoning} step is omitted for this task.

\subsubsection{Step 1: Retrieval}
\label{retrieval}
% 首先说
The retrieval module focus on gathering structural and semantic knowledge that may contribute to the completion of certain incomplete triple.
\paragraph{Supporting Triple Retrieval. }
In KGs, the attributes of an entity are represented in structural triples.  Different entities connected by the same relation often share common salient properties. These internal knowledge inherent in the graph structure provide the most direct support to the validity of a triple. Given an incomplete query triple in the form of $(h, r, ?)$ or $(?, r, t)$, we aim to retrieve $k$ supporting triples that are the most semantically similar to the incomplete query triple. Intuitively, we prioritize triples with the same entity and relation from the training set. If the number of available triples is less than $k$, we broaden our choices to triples with the same relation, and with entities similar to the known one in the query triple. 
\paragraph{Textual Context Retrieval.}
% 首先，要mention，triple和natural language sentence之间存在明显的semantic gap
% 短description和长description的作用区别
% 短description是激活LLM对某个实体的记忆，它肯定看过有关这个实体的知识
% 长的wiki document是限制LLM通过某个特别的语料来做提取
We note that there is a significant semantic gap between structural triples and natural language. For example, in Figure~\ref{KGCFigure}, entity ``Kasper Schmeichel'' is originally represented by entity id ``\textit{/m/07h1h5}'' while relation ``plays for sports teams'' is originally represented as \textit{``/sports/pro\_athlete/teams\allowbreak ./sports/sports\_team\allowbreak \_roster/team''}. Such a structured format is difficult for LLMs to process. 
%为了充分利用大语言模型的语义理解能力，我们从知识图谱所对应的知识库中提取和实体相关的context，并将raw representations of relations转化为通俗易懂的short phrases.
To fully leverage the semantic understanding capabilities of LLMs, we extract relevant contexts related to entities in the query triple and supporting triples from Wikidata knowledge base~\cite{wikidata}. % and convert the raw representations of relations into short phrases. 

In mainstream KGs, entities are represented in numerical or textual IDs. Each entity ID acts as an index to the data frame in its corresponding KB. Apart from triples, the data-frame of an entity contains significant contextual information such as entity label. To enhance data consistency across different KBs, identical entities across different KBs are aligned with the ``owl:sameAs'' property. Given its extensive coverage and up-to-date information, Wikidata is employed as our primary contextual corpus. Specifically, for each entity, we map the entity ID to Wikidata QID with the ``owl:sameAs'' property.~\footnote{Since Google Freebase is deprecated and migrated to Wikidata, we map the entity IDs in the FB15k237 dataset to corresponding Wikidata QIDs with official data dumps. } We then collect the textual entity label, the short description, and aliases from Wikidata URIs. Furthermore, Wikidata provides links to the associated Wikipedia pages of its entities. Considering the length of the document, we collect the first paragraph of these Wikipedia pages, which offer complementary semantic support for the completion of query triples. 
%Regarding the hierarchical relations in the FB15k237 dataset~\cite{FB15k237}, 

 % Therefore, we augment the two entities with their labels and descriptions for each selected triple. Regarding the hierarchical relations in FB15k237 dataset~\cite{Freebase}, we replace the raw representations with corresponding textual descriptions generated by~\citet{KICGPT}.

% In-context Learning的作用：1）告诉LLM你要做一个什么样的task，2）告诉LLM和这个当前triple有关的其他triple都有哪些，3）帮助LLM分析为什么结果是这个，给出一个合理的解释。
% 由于知识图谱高度结构化的特性，知识图谱补全并没有被用在LLM的预训练过程中，因此，我们需要为LLM赋予通过半结构化的语义知识中补全知识图谱三元组的能力。
% Due to the structured nature of KGs, KGC has not been included in the LLM pre-training paradigm. Therefore, in the first step, we conduct LLM in-context learning with three objectives: 1) introduce the KGC task to LLM, 2) provide related triples for demonstration, and 3) guide the LLM to reason across multiple triples. 
\paragraph{Candidate Answer Retrieval from KG. } 
% 第一第二句：做reranking的动机是我们不可能把kg中的所有entity都evaluate一遍，这太expensive，也不现实
% 第三第四句：我们在传统embedding-based KGC结果的基础上做re-ranking. 我们选前n个给大模型做reranking。
The widely adopted ranking-based evaluation for KGC task requires the model to score the plausibility of each entity in the KG as a potential replacement for the missing entity in the query triple. However, given the vast number of entities in the KG, employing LLMs to score and rank each entity is computationally expensive and impractical. 
Inspired by~\cite{RobustKGC,KICGPT,SSET}, we employ an embedding-based KGC model to initialize the scoring and ranking of entities within the KG. Here, we denote the ranked entity list as $\mathcal{A}_{\text{KGE}} = [e_1^{(k)}, e_2^{(k)}, ..., e_{n}^{(k)}, ..., e_{|\mathcal{E}|}^{(k)}]$, where the scoring function $f_r$ ensures a descendent ranking order. Formally, we have $f_r(h, e_i^{(k)}) < f_r(h, e_j^{(k)})$ if and only if $i > j$.  

\paragraph{Candidate Answer Retrieval from Text. } 
% 在第二阶段，我们将query entity的维基百科页面，需要做补全的query, 有关KGC任务的提示词，以及该query翻译过后的问题传给大语言模型。利用维基百科页面中的丰富上下文信息,大语言模型将可以运用其强大的信息提取和语义理解能力生成一些可能的答案候选。这里，我们把这些候选答案记作$\mathcal{A}_{\text{LLM}} = \{a_1, a_2, ..., a_m\}$, 它们为长尾实体提供重要的的语义信息。然而，生成式LLM并不保证输出符合特定规范的结果。 具体而言，输出的answer可能是KG中某个实体的alias，或者并不是KG中记录的实体，因此我们需要做进一步的后处理。
% 除了三元组，实体的维基百科页面也提供了丰富的语义信息。然而，受限于LLM的输入长度，我们不能提供完整的维基百科页面。不过，维基百科
Apart from supporting triples, the Wikipedia page of the known entity also entails rich semantic knowledge. %These semantic knowledge is crucially important for entities with a few neighbors.
% We aim to exploit the information extraction and semantic understanding capabilities of LLMs to generate potential answers through specific contexts of entities. 
Different from the short Wikidata description, the first Wikipedia paragraph provides a brief introduction to the entity. 
% 我们期待LLM可以通过更全面地了解一个实体，而
We anticipate that LLMs can harness their information extraction and comprehension capabilities by utilizing comprehensive contextual information about the known entity, thereby generating potential answers. Specifically, we pass the Wikipedia paragraph of the known entity and the natural language question translated from the query triple to the LLM. Based on the task-specific prompts, the LLM will output a list of answers in its response. However, it should be noted that generative LLMs do not guarantee that output answers will conform to entities in the KG. Therefore, we post-process the LLM output by replacing entity aliases with entity labels and filtering out invalid and unreliable answers that do not appear within the top-$\delta$ positions of $\mathcal{A}_{\text{emb}}$. Finally, we obtain a list of $m$ answers $\mathcal{A}_{\text{LLM}} = [e_1^{(l)}, e_2^{(l)}, ..., e_m^{(l)}]$, where $e_1^{(l)}, e_2^{(l)}, ..., e_m^{(l)} \in \mathcal{E}$, each of which is simultaneously supported by the LLM and the embedding model. 
\subsubsection{Step 2: Ranking}
\label{reranking}

% 别删， 这部分要放到related works里去。
% Existing KGC methods~\cite{TransE,RotatE,CompoundE} mainly focus on embedding entities and relations in the KG with pre-defined message-passing and scoring functions. However, due to information scarcity, KGs exhibit a pronounced long-tail characteristic~\cite{longtail,CLMKE,KICGPT}, where many entities appear infrequently. As a result, KG embeddings often fail to accurately represent all the dominant attributes of entities, especially those with few neighbors. This greatly affects the effectiveness of the widely used KGC paradigm. It is worth noting that several works~\cite{kgbert,KEPLER,SimKGC} attempt to encode entity embeddings with entity names and descriptions with language models. Nevertheless, despite undergoing expensive fine-tuning processes, these text-based KGC approaches still cannot outperform traditional embedding methods~\cite{CompoundE}. Inspired by the reasoning capability of LLMs~\cite{KICGPT}, we demonstrate a new KGC paradigm that integrates and utilizes contextual semantics from three different perspectives. 
% 首先，我们明确一个从SSET借鉴过来的观点：一个好的triple会同时受到语义知识和结构化知识的支持。
% 近期研究表明，entity的初始排序可能会影响大模型的排序性能。为了降低排序难度，我们收集同时被LLM和KGE模型选为候选答案的entity，并将它们重排到\mathcal{A}_{KGE}较前的位置。KICGPT4.3第二段有reference
Motivated by the complementary nature of semantic and structural knowledge, we aim to exploit the candidate answer list generated by the LLM and the KGE model to compose our rankings. %Recent studies~\cite{RankGPT,KICGPT} indicate that the initial order of items may significantly affect the performance of LLMs in ranking tasks. Hence,
To guide the LLM in utilizing entity descriptions for ranking candidate answers to query triples, we introduce supervised fine-tuning~(SFT) with LoRA adaptation~\cite{LLMRanker}. The training objective of SFT is to restore the original plausibility-based ranking for a list of shuffled candidate answers. Specifically, we construct training samples by corrupting the tail (or head) entity of each triple in the validation set. For each corrupted triple, we utilize an embedding-based model to initialize a ranked entity list and collect the top-$n$ entities as candidate answers. Then, we add the ground truth entity to the front of the candidate answer list, and shuffle the list randomly. %The shuffling process allows the LLM to be compatible with distinct embedding models. 
After that, we translate the masked triple to a question, and retrieve the entity label and the short Wikidata description for each candidate answer. Finally, we provide these questions along with their candidate answers and descriptions to the LLM for training. The LLM will learn to rank the candidate answers based on their contextual relevance and plausibility by considering the semantics of the question and entity descriptions.
% Finally, we provide the question to the LLM, and ask it to rank these candidate answers based on the , and fine-tune the LLM to restore the original order of the $\delta+1$ entities.  

During the inference stage, we construct a candidate answer set $\mathcal{C}$ with top-$n$ entities from $\mathcal{A}_{\text{KGE}}$ and all entities in $\mathcal{A}_{\text{LLM}}$. Formally, we have $\mathcal{C} = \allowbreak\mathcal{A}_{\text{KGE}}[0:n] \cup \mathcal{A}_{\text{LLM}}$. Then we employ the fine-tuned LLM to re-rank entities in $\mathcal{C}$ with their descriptions and the LLM's intrinsic knowledge. Subsequently, the LLM will output a re-ordered answer list $\mathcal{A}_{\text{RR}}=[\allowbreak e_1^{(o)}, \allowbreak e_2^{(o)}, ..., e_{|\mathcal{C}|}^{(o)}]$. Finally, we remove all entities in $\mathcal{C}$ from the original entity list $\mathcal{A}_{\text{KGE}}$, and compose the final ranking of all entities by attaching $\{\mathcal{A}_{\text{KGE}} \setminus \mathcal{C}\}$ to the end of $\mathcal{A}_{\text{RR}}$. 

\subsection{Knowledge Base Question Answering}
\begin{figure*}
    \centering
    \includegraphics[width=\textwidth]{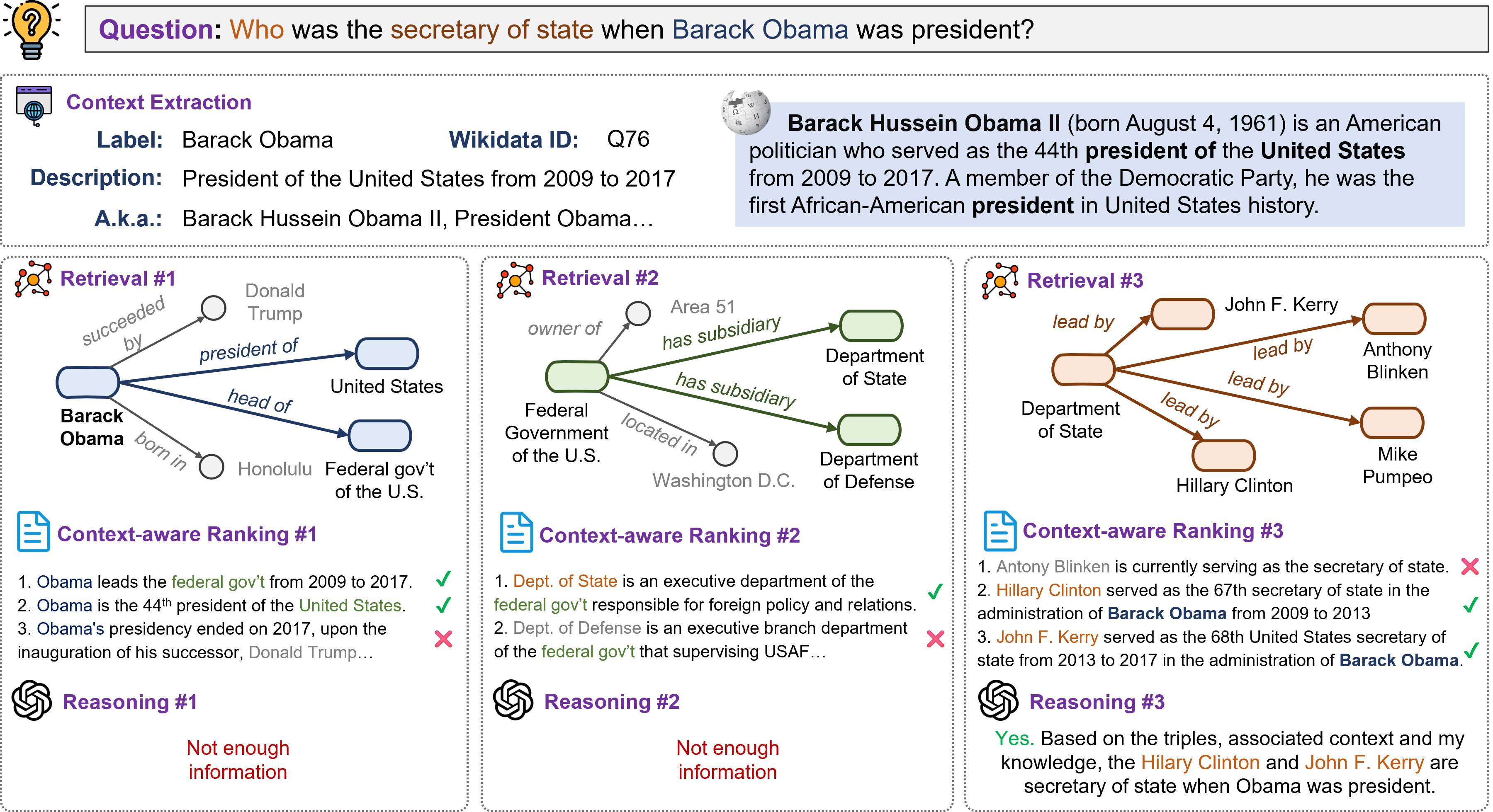}
    \caption{Knowledge Base Question Answering. }
    \label{KBQAFigure}
\end{figure*}
% Contextual ToG三步曲: 1) 在当前邻域中找到合适的path，2) 为path上的entity和triple augment语义信息，3) 检查当前邻域中的path和contextual信息是否足以回答问题，若不足以，则剪枝并扩大retrieval区域，若可以，则输出答案。
In this section, we introduce an in-context learning paradigm for the KBQA task (see Figure~\ref{KBQAFigure}). This paradigm focuses on the integration of contextual information, which plays a pivotal role in identifying plausible reasoning paths and facilitating the derivation of final answers.

Given a question $q$, we first identifies a set of $k$ topic entities $E^{(0)} = \{e_i^{(0)}\}_{i=1}^k$ with an LLM. 
% 這裏我們比ToG多的是associated contexts
Starting from these topic entities, we iteratively explore plausible reasoning paths until the LLM determines that it can answer the question based on the support of triples along the paths and their associated contexts. 
% 在整个过程中我们要维持和更新两类信息, 一个树状结构的path池，全局最相關句子池
Therefore, during the inference process, we maintain and update a set of reasoning paths $P=\{p_1, p_2, ..., p_M\}$ alongside a list of relation context sentences $C=\{rc_{1},rc_{2},..., rc_{N}\}$. Here, $M$ represents the width of the beam search, while $N$ denotes the number of relation context sentences. Each iteration of the process consists of three steps: 1) knowledge exploration, 2) reasoning path pruning, and 3) context-aware reasoning. 

At the beginning of the $D$-th iteration, each reasoning path consists of $D-1$ triples, i.e., $p_i = \{(h_n^{(d)}, r_n^{(d)}, t_n^{(d)})\}_{d=1}^{D-1}$, where $h_n^{(1)}$ is a topic entity from $E^{(0)}$, $t_n^{(d)} = h_n^{(d+1)}$ ensures the tail entity of one triple becomes the head entity of the next.~\footnote{WLOG, We only look for paths with forward relations. For each triple $(h, r, t)$, we introduce a reversed relation $r^{-1}$ and the reversed triple $(t, r^{-1}, h)$ into the KG. }

% Notes (by Muzhi)：我们Contextual ToG的运作时围绕着triple进行的。因此，我们这里的目的不是写什么relation search, relation prune, entity search, entity prune。上述这些是ToG的贡献，ToG开山鼻祖地提出了这一套RAG的框架，这里我选择避开这种写法，以突出我们的contribution （如下）
% 我们这里关注于知识在reasoning path上的传播，我们先 1）走一步，找到D-hop neighborhood中的triple，然后再2）给这些triple augment上context（即句子），并筛选出合适的triple，最后3）利用这些triple和句子解答问题。 
\subsubsection{Step 1: Context-aware Triple Retrieval}
In the initial step, we aim to retrieve candidate triples that can extend the reasoning paths. Specifically, for each reasoning path $p_m \in P$, we collect the tail entity $e_m^{(D-1)}$ from the last triple and identify the set of relations $R_m^{(D)}$ linked to the entity. 
We then construct queries in the form of $(e_m^{(D-1)}, r_m^{(D)}, ?)$ using each of the relations . Given that an entity can be linked to multiple relations, this process potentially increases the number of reasoning paths. To reduce the computational complexity, we exploit the LLM to select top-$M$ queries based on their relevance to the question. Subsequently, we proceed to complete the query triples by retrieving suitable neighboring entities from the KG, each of which derives a candidate triple that can potentially lead to answering the question. 

\subsubsection{Step 2: Candidate Entity Ranking}
In the second step, we focus on identifying those triples that are most likely to contribute to a correct answer. First, we augment each candidate triple with $\gamma$ relation context sentences that are best aligned with its contextual semantics as described in Section~\ref{ContextExtraction}. With relation contexts, we then exploit the LLM to select out top-$M$ triples from the candidate triples derived from each query $(e_m^{(D-1)}, r_m^{(D)}, ?)$. This helps us to prune out irrelevant and noisy neighboring entities that could mislead the LLM into producing incorrect answers. Due to the length limit of LLM inputs, it is still impractical to leverage the remaining $M\times M$ triples in knowledge reasoning. Therefore, we further refine our selection from the remaining triples to top-$M$ triples with the highest contextual relevance between the relation contexts and the question~\footnote{We utilize the bge-large-en-v1.5 model to measure the semantic similarity of the question and each supporting sentence.}. Finally, we attach the $M$ triples to the end of each corresponding reasoning path and append their relation contexts into the context list $C$. The context list $C$ are then updated by ranking their relevance to the given question and only top-$N$ relation context sentences are remained at the end of this step.

\subsubsection{Step 3: Context-aware Reasonin}
Upon obtaining the new top-$M$ reasoning paths $P$ and updating relation context list $C$, this extra knowledge retrieved from the CG are integrated into the origin question as a part of the prompt. The prompt is input to the LLM and the LLM perform the reasoning step to determine whether the sufficient information has been retrieved from the CG. If yes, the LLM generates the answer based on the retrieved knowledge and its inherent knowledge. Otherwise, the whole process will iterate by starting the first step with new reasoning paths $P$ and relation context set $C$.
% wheather the extra knowledge retrieved from D-hop neighborhood.
\section{Experiments on KG Completion}
In this section, we assess the effectiveness of $\text{KGR}^3$ in the KGC task. Our investigation is guided by the three following research questions:
\begin{itemize}[leftmargin=1em]
    \setlength\itemsep{0.1em}
    % \item We propose a novel ``\textit{retrieval, reasoning, re-ranking}'' framework $\text{KGR}^3$ for the KGC task, ...
    \item \textbf{RQ1:} Whether $\text{KGR}^3$ works for varied embedding methods? 

    \item \textbf{RQ2:} Whether different types of entity contexts contribute to enhancing knowledge reasoning?
   
    \item \textbf{RQ3:} Can LLM effectively leverage entity contexts for the KGC task with or without SFT? 

    \item \textbf{RQ4:} Can CGR$^3$ improve the inference performance for predicting long-tail entities?
    %significantly outperforms the existing state-of-the-art approaches.
\end{itemize}
\subsection{Datasets}
% We utilize two widely-used datasets, namely FB15k237~\cite{FB15k237} and WN18RR~\cite{ConvE} to evaluate our proposed method. FB15k237 is derived from Freebase~\cite{Freebase}, an encyclopedic KG containing general knowledge about topics such as celebrities, organizations, movies and sports. WN18RR is a subset of WordNet~\cite{WordNet}, a lexical KG with knowledge about English morphology. To prevent potential data leakage, FB15k237 and WN18RR excludes reversible relations from their backend KGs. Detailed statistics of datasets are shown in Table~\ref{}.
We evaluate our proposed framework on two widely-used datasets FB15k237~\cite{FB15k237} and YAGO3-10~\cite{YAGO3-10}. FB15k237 is derived from Freebase~\cite{Freebase}, an encyclopedic knowledge base containing general knowledge about topics such as celebrities, organizations, movies, and sports. YAGO3-10 is a subset of YAGO3~\cite{YAGO3-10}, a knowledge base built upon Wikipedia, WordNet~\cite{WordNet}, and GeoNames~\cite{GeoNames}. To prevent potential data leakage, FB15k237 excludes reversible relations from the backend KB. Detailed statistics of the two datasets are shown in Table~\ref{dataset}.
% We utilize two widely-used datasets, namely FB15k237~\cite{FB15k237} and YAGO3-10~\cite{YAGO} to evaluate our proposed method. FB15k237 is derived from Freebase~\cite{Freebase}, an encyclopedic KG containing general knowledge about topics such as celebrities, organizations, movies and sports. WN18RR is a subset of WordNet~\cite{WordNet}, a lexical KG with knowledge about English morphology. To prevent potential data leakage, FB15k237 and WN18RR excludes reversible relations from their backend KGs. Detailed statistics of datasets are shown in Table~\ref{}.

\begin{table}[htbp]\centering
\small
% \resizebox{0.87\linewidth}{!}{
\begin{tabular}{ccc}
\toprule
\textbf{Dataset}& \textbf{FB15k237} & \textbf{YAGO3-10}\\
\midrule
\#Entities & 14,541 & 123,182 \\
\#Relations & 237 & 37 \\
\#Train & 272,115 & 1,079,040 \\
\#Valid & 17,535 & 5,000 \\
\#Test & 20,466 & 5,000 \\
\bottomrule
\end{tabular}
% }
\caption{Statistics of Datasets}
\label{dataset}
\vspace{-0.3cm}
\end{table}

% KGR2是没有reasoning
\begin{table*}[t]
\centering
\small
% \resizebox{0.9\textwidth}{!}{
\begin{tabular}{lccccccccc}
\toprule
\multirow{2}{*}{\textbf{Model}} & \multicolumn{4}{c}{\textbf{FB15K-237}} & \multicolumn{4}{c}{\textbf{YAGO3-10}} \\
\cmidrule(lr){2-5} \cmidrule(lr){6-9}
& MRR & Hits@1 & Hits@3 & Hits@10 & MRR & Hits@1 & Hits@3 & Hits@10 \\
\midrule

% ComplEx & 0.273 & 0.197 & 0.298 & 0.425 & 0.379 & 0.291 & 0.426 & 0.546 \\
ComplEx & 0.247 & 0.158 & 0.275 & 0.428 & 0.360 & 0.260 & 0.400 & 0.550 \\ % 原始指标
ComplEx + $\text{KGR}^2$ & 0.315 & 0.248 & 0.343 & 0.428 & 0.402 & 0.336 & 0.430 & 0.537 \\
ComplEx + $\text{KGR}^3$ & \textbf{0.333} & \textbf{0.263} & \textbf{0.365} & \textbf{0.460} & \textbf{0.408} & \textbf{0.340} &\textbf{0.441} & \textbf{0.559} \\
Improvements & 34.82\% & 66.46\% & 32.73\% & 7.48\% & 13.33\% & 30.77\% & 10.25\% & 1.64\% \\
\midrule
% RotatE & 0.348 & 0.253 & 0.383 & 0.540 & 0.493 & & & \\
RotatE & 0.338 & 0.241 & 0.375 & 0.533 & 0.495 & 0.402 & 0.550 & 0.670 \\ % 原始指标
RotatE + $\text{KGR}^2$ & 0.370 & 0.283 & 0.404 & 0.542 & 0.508 & 0.422 & 0.553 & 0.662 \\

RotatE + $\text{KGR}^3$ & \textbf{0.382} & \textbf{0.293} & \textbf{0.417} & \textbf{0.559} & \textbf{0.521} & \textbf{0.443} & \textbf{0.572} & \textbf{0.678} \\
Improvements & 13.02\% & 21.58\% & 11.20\% & 4.88\% & 5.25\% & 10.20\% & 4.00\% & 1.19\% \\
\midrule
% GIE & 0.363 & 0.269 & 0.399 & 0.554 &  &   &  &  \\
GIE & 0.362 & 0.271 & 0.401 & 0.552 & 0.579 & 0.505 & 0.618 & \textbf{0.709} \\ % 原始指标
GIE + $\text{KGR}^2$ & 0.378 & 0.288 & 0.412 & 0.557 & \textbf{0.599} & \textbf{0.522} & \textbf{0.633} & 0.702\\
GIE + $\text{KGR}^3$ & \textbf{0.391} & \textbf{0.301} & \textbf{0.426} & \textbf{0.573} & 0.597 & 0.518 & 0.625 & 0.698 \\
Improvements & 8.01\% & 11.07\% & 6.23\% & 3.80\% & 3.45\% & 3.37\% & 2.43\% & -0.99\% \\
\midrule
Avg. Improvements & 18.62\% & 33.04\% & 16.72\% & 5.39\% & 7.34\% & 14.78\% & 5.56\% & 0.61\% \\
\bottomrule
\end{tabular}
% }
\caption{Experiment results of the KGC task on FB15k-237 and YAGO3-10 datasets. The best results are in \textbf{bold}. }% Note: $\text{KGR}^2$ is a simplified variant of $\text{KGR}^3$ without ``\textit{reasoning}'' module. }
\label{tab:results}
\vspace{-0.0cm}
\end{table*}

\subsection{Baselines}
In this section, we evaluate the efficacy of our proposed $\text{KGR}^3$ framework by integrating it with three widely utilized embedding-based KGC models: RotatE~\cite{RotatE}, ComplEx~\cite{Complex}, and GIE~\cite{GIE}. These models not only serve as baseline methods but are also foundational for candidate answer retrieval. Instead of surpassing all baseline methods, our main objective is to evaluate the effectiveness of our context-enriched $\text{KGR}^3$ framework when applying to different embedding models. Hence, we deliberately include a limited selection of baseline models. %In Table~\ref{tab:results}, 

\subsection{Implementation Details}~\label{sec:implementation}
We conduct all of our experiments on a Linux server with two Intel Xeon Platinum 8358 proces-
sors and eight A100-SXM4-40GB GPUs. We choose the framework provided by the GIE~\cite{GIE} project for training the base embedding models, strictly following the parameter settings provided. During the reasoning stage, we utilize OpenAI's gpt-3.5-turbo-0125 checkpoint~\footnote{https://platform.openai.com/docs/models}. The Re-ranking stage employs Meta-Llama-3-8B-Instruct with BF16 precision as the backbone model~\footnote{https://huggingface.co/meta-llama/Meta-Llama-3-8B-Instruct}. The SFT task is implemented based on the LLaMA-Factory~\cite{llamafactory} framework and applies LoRA technique~\cite{lora}, with a rank setting to 16 and an alpha setting to 32. Additionally, AdamW~\cite{adamW} is used as the optimizer, the batch size is set to 2 per device, the gradient accumulation steps is set to 4, and the learning rate is 1.0e-4. The sampling ratio of the validation set is 5\%, and the best checkpoint is selected based on evaluation loss.

\subsection{Evaluation} 
For each query triple in the form of $(h, r, ?)$ or $(?, r, t)$, the KGC model outputs a ranked list of all entities in the KG. For a fair comparison, we adopt the ``filtered'' setting introduced in~\cite{TransE}. Except for the ground truth entity, we remove all other valid entities that conform to an existing triple in training, validation, or test set from the ranked list in advance. Based on the position of the ground truth entity, We compute Hits@$1$, Hits@$3$, Hits@$10$ and mean reciprocal rank~(MRR), where higher results indicate better performance.

\subsection{Main Results}
% Main Results 4个重点
% 1. 整体实验效果好，提升了多少
% 2. Hits@1比Hits@10提升多
% 3. 性能越差的模型提升越大
% 3. KGR^2 和 KGR^3的对比：Triple中不足够的知识可以通过长文本补全，有效提升 Hit@10，突破base embedding模型的界限
Table~\ref{tab:results} summarizes the performance of the $\text{KGR}^3$ framework on three different base embedding methods. The experiment results show that $\text{KGR}^3$ and its simplified variants $\text{KGR}^2$ without ``\textit{reasoning}'' module significantly and consistently enhances each embedding method among all metrics. On average, our $\text{KGR}^3$ framework improves the Hits@$1$ by 33.04\% and 14.78\% on FB15k-237 and YAGO3-10 datasets. These results demonstrate the effectiveness and superiority of integrating LLMs and entity contexts with embedding-based KGC models, which address our \textbf{RQ1}. 

Notably, the improvement in Hits@1 is more substantial than that in Hits@3 and Hits@10. This indicates that the $\text{KGR}^3$ framework is particularly effective at identifying the most accurate answers. Since our framework primarily focuses on re-ordering top-$n$ (or top-$\delta$ if we consider reasoning outputs) entities from the initial ranked entity list, the upper bound of Hit@1, Hit@3, and Hit@10 are implicitly constrained by the Hits@$n$ or Hit@$\delta$ performance of the base embedding model. Given that Hits@1 is typically further from this upper bound, the potential for improvement will be greater. Additionally, by leveraging semantic knowledge from entity contexts, the LLM gains a more comprehensive understanding of the entities, thereby enabling more precise inferences, particularly for top-ranked candidate answers. 

% 接下来一段逻辑这么来：
% Furthermore, the performance gains are more pronounced for simple embedding models.  简单的embedding模型可能不能充分捕捉图谱的结构信息，因此会在candidate list中引入更多的noise (或 noisy entities 选一个说法)。 在提供了candidate answer的description之后，LLM就能利用它的语义理解能力，把这个list中不符合query triple语义的entity识别出来放到后边去。就这样，KGR3间接地提升了embedding based model的鲁棒性。 
Furthermore, the performance gains are more pronounced for simpler embedding models such as ComplEx~\cite{Complex}. Simple embedding models cannot fully capture the structural information in the KG, leading to the introduction of noisy entities in the candidate answer list. With entity descriptions, the LLM can utilize its semantic understanding capabilities to identify and deprioritize candidate answers that do not match the semantics of the query triple. Hence, $\text{KGR}^3$ can enhance the robustness of these embedding models. 
% This argument can also be supported by the experimental results in Appendix~\ref{sec:appendixA}, which shows that $\text{KGR}^3$ can significantly improve the ability of KG embedding models to predict long-tail entities.

% Finally, a comparison between $\text{KGR}^2$ and $\text{KGR}^3$ reveals that the ``\textit{reasoning}'' module provides a substantial boost in performance. 有些时候，KG is not able to provide sufficient structural knowledge to derive plausible answers。 在这种情况下, the long Wikipedia paragraph can effectively enrich the semantic knowledge about certain entity. 而LLM可以通过external semantic knowledge提供额外的candidate answer. 这surpass the inherent limitations of KG本身，因而我们看到了显著的性能提升，尤其是在hit@10上。
In addition, a comparison between $\text{KGR}^2$ and $\text{KGR}^3$ reveals that the inclusion of the ``\textit{reasoning}'' provides a notable boost. In certain scenarios, the KG may lack sufficient structural information to derive plausible answers. Nevertheless, long Wikipedia paragraphs can effectively augment specific entities with extra semantic knowledge, which allows the LLM to generate additional candidate answers with its semantic reasoning capability. This surpasses the inherent limitations of KGs, leading to substantial performance improvements. An case study showing the effectiveness of the \textbf{Reasoning} and \textbf{Re-ranking} processes is demontrated in Appendix~\ref{sec:appendix1}-\ref{sec:appendix3}

\subsection{Ablation Studies}
\subsubsection{Effectiveness of Entity Contexts}
To address RQ2, we assess the contribution of different types of contexts in the reasoning and re-ranking modules of $\text{KGR}^3$, and conduct ablation studies on FB15k-237 dataset. 
% 主要写四段
% 第一段写$\text{KGR}^3$ w/o context in Reasoning. replace the entity description and wikipedia document with entity name. 这时，LLM只能通过entity label来检索内部的知识，并生成可能的答案。显著的性能下降显示
% 第二段写$\text{KGR}^3$ w/o context in Reasoning. 在这个对比实验中我们简单地remove entity description，造成了显著的性能下降。我们认为性能下降是因为SFT依赖query triple和candidate entities之间的语义相关性来进行ranking，
% 第三段写，当我们去除了所有context的时候，性能进一步下降。这时候我们实际上运用的是大语言模型的信息检索能力，而非语意理解能力，而前者并不是大语言模型所擅长的
% 第四段写，在有contextual information的帮助下，即使是基于一个较弱的base embedding model，KGR3仍旧可以surpass state-of-the-art embedding based and text based KGC model. 这佐证了KG中semantic knowledge and structural knowledge的互补性，为未来的KG研究指明了方向。
In the ``$\text{KGR}^3$ w/o context in Reasoning'' variant, we remove the short descriptions used to explain the entities and replace the Wikipedia paragraph of the known entity with an entity label. Under such circumstances, the LLM cannot fully demonstrate its strong semantic understanding capability, resulting in lower performance.
%Under such circumstances, the LLM relies solely on entity labels to evoke its internal knowledge about the entity. %The significant performance drop emphasizes that the internal knowledge is insufficient for the LLM to generate plausible answers for specific incomplete triple, which further hightlights the importance of utilizing entity contexts from the knowledge base. 

In the ``$\text{KGR}^3$ w/o context in Re-ranking'' variant, we simply remove the entity descriptions for each candidate answer, which results in a noticeable performance decline. This decline reveals that LLMs may lack a fundamental understanding of certain entities within the KG. Consequently, without sufficient semantic information, the LLM cannot rank candidate answers effectively.

If we remove all contextual information from the $\text{KGR}^3$ framework, performance deteriorates even further. This indicates that every type of context is meaningful and irreplaceable, playing a crucial role at each stage of the process. Without entity contexts, the LLM only relies on its inherent knowledge, hence leading to suboptimal inference results. 
%In this scenario, the LLM relies primarily on its information retrieval capabilities rather than its semantic understanding abilities. Since large language models are not particularly adept at open-domain information retrieval, 

% $\text{KGR}^3$ 超过了embedding-based sota CompoundE和text-based sota SimKGC with proper base embedding model. 这说明了 entity context 可以补足embedding method无法充分捕捉结构化知识的缺陷，重申了它的重要性。此外，SimKGC和KGR3的差距体现了原有text-based methods的局限性。一方面，以PLM驱动的模型语义理解能力不够强，即使经过finetune也无法取得好的性能。 另一方面，这些方法underutilize KG中的semantic and structural info。当用于补全某个具体的triple时，它们往往只考虑triple本身，而忽略了entity的local neighborhood，相似的其他triple.
With proper base embedding model, \text{KGR\textsuperscript{3}} surpasses the state-of-the-art embedding-based model CompoundE~\cite{CompoundE} and the text-based model SimKGC~\cite{SimKGC}. This demonstrates that entity context can compensate for the limitation of embedding methods in modelling the graph structure. Furthermore, discrepancies between SimKGC and \text{KGR\textsuperscript{3}} underscores the limitations of existing text-based methods. On the one hand, PLM-driven models exhibit insufficient semantic understanding, and the gap between lightweighted PLM and LLM cannot be easily alleviated by fine-tuning. On the other hand, these methods underutilize the semantic and structural information within KG. When being applied to complete a specific triple, they often consider the triple in isolation, neglecting the local neighborhood of the known entity and other similar triples.

% With the assistance of contextual information, even when based on a weaker base embedding model, $\text{KGR}^3$ can still surpass state-of-the-art embedding-based and text-based KGC models. This outcome underscores the complementary nature of semantic and structural knowledge within knowledge graphs, pointing the way for future research in this area.

\begin{table}
\centering
\resizebox{0.48\textwidth}{!}{
\begin{tabular}{lcccc}
\toprule
\textbf{Settings} & MRR & Hits@1 & Hits@3 & Hits@10 \\
\midrule
ComplEx + $\text{KGR}^3$ & \textbf{0.333} & \textbf{0.263} & \textbf{0.365} & \textbf{0.460} \\
- w/o context in \textbf{Reasoning} & 0.330 & 0.260 & 0.361 & 0.454 \\
- w/o context in \textbf{Re-ranking} & 0.319 & 0.245 & 0.351 & 0.453 \\
- w/o all contexts & 0.305 & 0.235 & 0.336 & 0.428 \\
\midrule
RotatE + $\text{KGR}^3$ & \textbf{0.382} & \textbf{0.293} & \textbf{0.417} & \textbf{0.559} \\
- w/o context in \textbf{Reasoning} & 0.375 & 0.285 & 0.411 & 0.555 \\
- w/o context in \textbf{Reranking}& 0.361 & 0.264 & 0.398 & 0.559 \\
- w/o all contexts & 0.360 & 0.262 & 0.398 & 0.561 \\
\midrule

GIE + $\text{KGR}^3$ & \textbf{0.391} & \textbf{0.301} & \textbf{0.426} & \textbf{0.573} \\
- w/o. context in \textbf{Reasoning} & 0.384 & 0.290 & 0.422 & 0.574 \\
- w/o. context in \textbf{Re-ranking} & 0.366 & 0.267 & 0.403 & 0.572 \\
- w/o. all contexts & 0.363 & 0.267 & 0.400 & 0.556 \\
\midrule
CompoundE & 0.357 & 0.264 & 0.393 & 0.545 \\
SimKGC & 0.336 & 0.249 & 0.362 & 0.511 \\
\bottomrule
\end{tabular}
}
\caption{Ablation Experiments on FB15k-237 dataset with different combinations of contexts.}
\label{tab:RQ2}
\end{table}

\subsubsection{Effectiveness of SFT} 
% In response to \textbf{RQ3}, we conduct extra experiments on $\text{KGR}^3$ with different LLMs. 
% 从table 4中可以看出当我们不使用SFT finetune re-ranking的时候，performance明显降低，甚至有可能低于embedding model的原始性能。这表明了 纵使pre-trained vanilla LLM有着很强的语义理解能力，但它并不会做re-ranking。这种根据entity context进行排序的能力是在fine-tune的过程中acquire到的。相较于Llama, ChatGPT具有更强的语义理解和指令跟随能力，因而achieve better performance. 然而, 由于结构化的三元组和naturallanguage sentence之间的语义差距，ChatGPT并无法超过finetuned Llama, 进一步显示了SFT的重要性。
\begin{table}
\centering
\resizebox{0.48\textwidth}{!}{
\begin{tabular}{lcccc}
\toprule
\textbf{Settings} & MRR & Hits@1 & Hits@3 & Hits@10 \\

\midrule
ComplEx + $\text{KGR}^3$ & \textbf{0.329} & \textbf{0.256} & \textbf{0.363} & \textbf{0.456} \\
% w/o SFT & & & & \\ % 这一行不用填
  - w/ non-SFT Llama3  & 0.288 & 0.206 & 0.323 & 0.450 \\
  - w/ ChatGPT & 0.299 & 0.224 & 0.330 & 0.453\\
  \midrule
RotatE + $\text{KGR}^3$ & \textbf{0.380} & \textbf{0.287} & \textbf{0.417}& \textbf{0.565} \\
% w/o SFT & & & & \\ % 这一行不用填 
  - w/ non-SFT Llama3  & 0.321 & 0.215 & 0.356 & 0.556 \\
  - w/ ChatGPT & 0.348 & 0.248 & 0.387 & 0.559 \\
\midrule
GIE + $\text{KGR}^3$ & \textbf{0.383} & \textbf{0.291} & \textbf{0.418} & \textbf{0.576} \\
%w/o SFT & & & & \\ % 这一行不用填
  - w/ non-SFT Llama3  & 0.324 & 0.213 & 0.364 & 0.564 \\
  - w/ ChatGPT & 0.354 & 0.253 & 0.391 & 0.570 \\
\midrule
KICGPT w/ limited demos & 0.274 & 0.183 & 0.280 & 0.496 \\
\bottomrule
\end{tabular}
}
\caption{The performance of $\text{KGR}^3$ without SFT on the first 2,000 examples of FB15k-237 dataset.}
\label{tab:RQ3}
\vspace{-0.2cm}
\end{table}
In response to \textbf{RQ3}, we conduct extra experiments on $\text{KGR}^3$ with different LLMs. From Table~\ref{tab:RQ3}, we observe that if we remove SFT step from the re-ranking module, the performance significantly decreases, even potentially falling below base embedding models. Despite with certain semantic understanding capabilities, vanilla LLMs cannot perform well in ranking tasks. We can further conclude that the ability to perform ranking based on entity context is acquired during the fine-tuning process. Compared to Llama, ChatGPT achieves a better performance with its stronger instruction following capability. Nevertheless, ChatGPT still lags far behind the finetuned Llama, showcasing the necessity of SFT. 

Moreover, we compare $\text{KGR}^3$ with state-of-the-art LLM-based KGC baseline KICGPT~\footnote{We only modify the parameters \texttt{demo\_per\_step} to 2, \texttt{max\_demo\_step} to 2 and \texttt{candidate\_num} to 10 in ~\cite{KICGPT}, to ensure the consistency with the settings in this work. Since there is no metric evaluation provided, we evaluated the natural language results generated within our framework.}. It should be noted that KICGPT processes all triples in the KG with the same entity or relation as the incomplete triple, which consumes far more ($20\times$) tokens than our $\text{KGR}^3$ framework. For a fair comparison, we re-evaluate KICGPT with $k$ supporting triples. From the experimental results we observe that KICGPT left significantly behind all variants of $\text{KGR}^3$. The remarkable performance gap can also be explained by the introduction of SFT since KICGPT employs ChatGPT as its backbone.  
% 这里要不要讲我们很高效，感觉讲这个就得把kicgpt原来的数据给出来。
% 那个不同k值的实验应该也没时间做了。

% Some recent research表明，item的初始顺序会影响LLM排序的最终结果。在这种情况下，LLM可能会以原始的candidate answer list作为shortcut，并输出答案，进而影响性能 这个理论不对啊！如果是这样的话GIE KGR3 w/o SFT最低也是和GIE 一样
%From Table~\ref{tab:RQ3} we observe 
\subsubsection{Effect on Handling Long-tail Entities}
\begin{figure}
    \vspace{-0.25cm}
    \centering
    \includegraphics[width=0.45\textwidth]{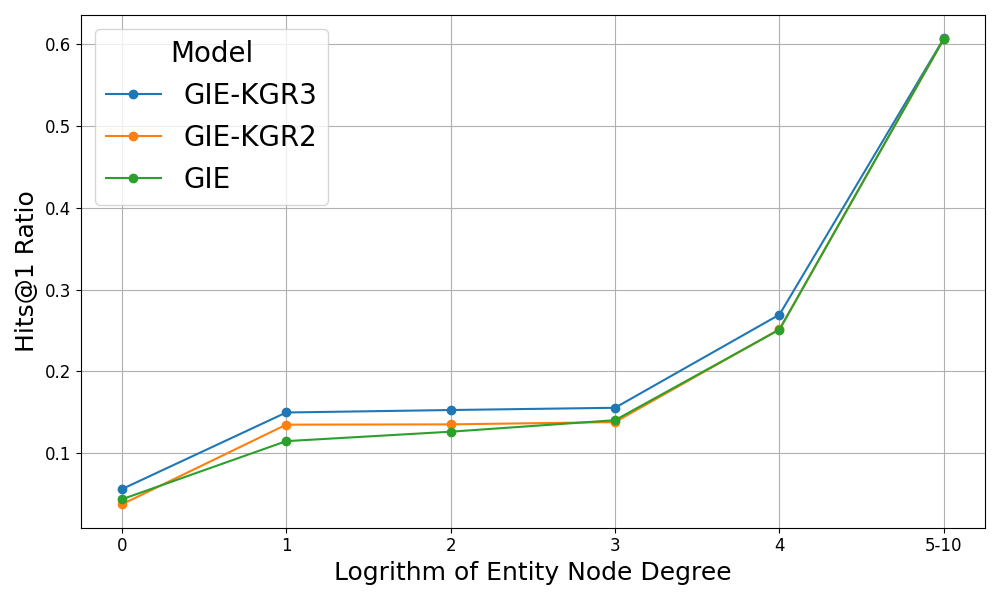}
    \caption{Average Hit@$1$ performance of GIE, GIE-$\text{KGR}^2$ and GIE-$\text{KGR}^3$ grouped by the logarithm of entity node degree on FB15k-237 dataset.}
    \label{fig:longtail}
\vspace{-0.08cm}
\end{figure}
In response to \textbf{RQ4}, we follow~\cite{ijcai2022p0318,KICGPT} and group triples from FB15k-237 test set into $5$ classes with the logarithm of the node degree of their known entities. We average the Hit@$1$ performance of each group of triples with $\text{KGR}^3$, $\text{KGR}^2$ (w/o reasoning module), and their base embedding model GIE~\cite{GIE} (see Figure~\ref{fig:longtail}). From Figure~\ref{fig:longtail} we observe that $\text{KGR}^3$ consistently outperforms its variant $\text{KGR}^2$ and GIE in all groups, especially for the first two groups where entities have fewer neighbors. This empirically shows that the proposed framework can effectively alleviate the long-tail problem. In addition, the performance gap between $\text{KGR}^2$ and GIE is less pronounced, which reaffirms the importance of the reasoning part, where the LLM generates possible answers based on the Wikipedia introduction of entities.  

\section{Experiments on KGQA}
\subsection{Datasets and Evaluation Metric}
We note that a lot of commonly-used KGQA benchmarks like CWQ~\cite{cwq_dataset} and WebQSP~\cite{webqsp_dataset} are constructed from Freebase~\cite{Freebase} which has been defunct since 2015. Some of the knowledge in Freebase is outdated or contradicts information in Wikipedia~\cite{WWQ}. Clearly, compared to Freebase, the knowledge in Wikipedia has higher coverage and accuracy, and in this work, Wikipedia serves as the main source of contextual information. Our assumption is that the contextual information can support or complement the triple-based knowledge in the KG, rather than contradict it. Therefore, we  consider KGQA datasets based on Wikidata where the triple-based knowledge is better aligned with the contextual information from Wikipedia, rather than KGQA datasets constructed from Freebase. 

In this work, QALD10-en~\cite{qald10} and WikiWebQuestion (WWQ)~\cite{WWQ} are used as KGQA datasets for evaluation. QALD10-en is a new, complex, Wikidata-based KGQA benchmarking dataset as the 10th part of the Question Answering over Linked Data (QALD) benchmark series.
WWQ is constructed by migrating the popular WebQSP~\cite{webqsp_dataset} benchmark from Freebase to Wikidata, with updated SPARQL and up-to-date answers from the much larger Wikidata.

For all datasets, exact match accuracy (EM) is used as our evaluation metric following previous works \citep{cok, ToG}.
% \section{Related Works}
% \paragraph{Text-rich Knowledge Graph } 
% \paragraph{LLM-based Knowledge Graph Reasoning } 

\subsection{Baseline}
We compare with standard prompting (IO prompt) \citep{io_prompt}, Chain-of-Thought prompting (CoT prompt) \citep{COT}, and Self-Consistency \citep{self_consistency} with 6 in-context exemplars and "step-by-step" reasoning chains. Moreover, for each dataset, we pick previous state-of-the-art (SOTA) works for comparison. We notice that fine-tuning methods trained specifically on evaluated datasets usually have an advantage by nature over methods based on prompting without training, but sacrificing the flexibility and generalization on other data. Therefore, we compare with previous SOTA among all prompting-based methods and previous SOTA among all fine-tuned (FT) methods respectively. With regard to previous prompting-based methods, we select their results achieved with GPT-3.5 for a a fair play.

\subsection{Implementation}
We use ChatGPT (GPT-3.5-turbo) as the backbone LLM for CGR$^3$ by calling OpenAI API. The maximum token length for the generation is set to 256. In all experiments, we set both width $M$ and depth $D_{max}$ to 3 for beam search. We use 5 shots in CGR$^3$-reasoning prompts for all the datasets.

\subsection{Experimental Results}
\begin{table}[t!]
\centering
\resizebox{.48\textwidth}{!}{
\begin{tabular}{lcc}
\hline
\multirow{2}{*}{\textbf{Method}} & \multirow{2}{*}{\textbf{QALD10-en}} & \multirow{2}{*}{\textbf{WWQ}} \\
                                 &                               &                              \\ \hline
\multicolumn{3}{c}{\textit{Without external knowledge}}   \\ \hline
IO prompt w/ChatGPT  &   42.0    &    57.7   \\
SC w/ChatGPT  & 42.9 & - \\ 
SC w/ChatGPT   &   45.3    &    -         \\ \hline
\multicolumn{3}{c}{\textit{With external knolwedge}}   \\ \hline
Prior FT SOTA  &   45.4$^\alpha$    &    65.5$^\beta$         \\
Prior Prompting SOTA  &   50.2$^\theta$   &    72.6$^\theta$          \\
 \hline
\multicolumn{3}{c}{\textit{Ours}}   \\ \hline
CGR$^3$                      &  \textbf{54.7}     &  \textbf{78.8}           \\
CGR$^3$ w.o./Context                  &38.1   &  67.3            \\
\texttt{Gain}  &  \textcolor[RGB]{0,128,0}{(+43.6)}      &  \textcolor[RGB]{0,128,0}{(+17.1)}             \\ \hline
\end{tabular}
}
\caption{Exact match accuracy of CGR$^3$ using ChatGPT as  the backbone models on QALD10-en and WWQ. The prior FT (Fine-tuned) and prompting SOTA include the best-known results: $\alpha$: \cite{qald10-en1}; $\beta$: \cite{WWQ}; $\theta$: \citet{ToG}}.
\label{tab: kbqa}
\end{table}
Since CGR$^3$ uses external KGs and contextual information to enhance LLM, we first compare it with those methods leveraging external knowledge as well. As we can see in Table \ref{tab: kbqa}, even if CGR$^3$ is a training-free prompting-based method and has natural disadvantage in comparison with those fine-tuning methods trained with data for evaluation, CGR$^3$ still achieves new SOTA performance in both datasets.  If comparing with other promoting-based methods with ChatGPT as backbone models (especially ToG), CGR$^3$ can win the competition on all datasets. 

It is noteworthy that other prompting-based methods rely solely on triple knowledge from KGs, whereas CGR$^3$ allows the LLM to leverage additional contextual information for more precise reasoning on KGs. This is likely the primary reason why CGR$^3$ outperforms other prompting-based methods. To verify this, we evaluated a variant of CGR$^3$ that excludes contextual information for comparison. As shown in Table \ref{tab: kbqa}, incorporating contextual information results in a relative increase of 43.6\% and 17.1\% in Exact Match (EM) on QALD10-en and WWQ, respectively. These experimental results support our hypothesis that KGQA methods can significantly benefit from the integration of contextual information.

\section{Conclusion}
This work points out several critical shortcomings of triple-based KGs, including their inability to represent diverse knowledge flexibly and perform complex knowledge reasoning accurately, due to the lack of contextual information. By highlighting these limitations, we underscore the necessity of moving beyond triple-based representation for KGs and introduce the concept of CGs. CGs integrate rich contextual data, such as temporal, geographic, and provenance information, thus providing a more comprehensive and accurate representation of knowledge. This enhanced representation supports more effective reasoning by leveraging the added layers of contextual information.

To verify the effectiveness of incorporating contexts on knowledge representation and reasoning, we present CGR$^3$, a novel knowledge reasoning paradigm that integrates LLMs (LLMs) with CGs to address the limitations of traditional triple-based knowledge reasoning methods.  Through extensive experiments on KG completion and KG question answering tasks, we demonstrated that incorporating contextual information significantly improves the performance of existing models. Our results underscore the importance of context in capturing the complexity and richness of real-world knowledge, enabling more nuanced and accurate inferences.

In conclusion, the introduction of CGs represents a significant step forward in the evolution of KGs, offering a more sophisticated and comprehensive approach to knowledge representation and reasoning. This work opens new avenues for future research and applications, highlighting the potential of CGs and LLMs in advancing the field of artificial intelligence.
% \section*{Limitations}

\bibliography{acl_latex}

\clearpage

\appendix
\onecolumn
\section{Appendix}

\subsection{Prompt templates of Retrieval stage}\label{sec:appendix1}
Table~\ref{tab:pt_retrieval} shows the prompt templates of the \textbf{Retrieval} stage and give an example from FB15k237.
\begin{table}[h!]
    \centering
    \begin{tabular}{p{\textwidth}} % Change to full width
        \toprule
        \textcolor{darkblue}{\textbf{\#\# KG Triplet for completion}:}
        ([MASK], \textit{/location/adjoining\_relationship/adjoins}, Champaign) \\
        \\
        \textcolor{darkblue}{\textbf{\#\# Task for completion}:}
        "The question is to predict the head entity [MASK] from the given ([MASK], \textit{location adjoining\_relationship adjoins}, Champaign) by completing the sentence 'Champaign is the adjoins of what location? The answer is '."\\
        \midrule
        \textcolor{darkblue}{\textbf{\#\# Task demonstrations}}: \\
        \\
        \#\# \textbf{Demo 1}: "The question is to predict the head entity [MASK] from the given ([MASK], \textit{location adjoining\_relationship adjoins}, Washington County) by completing the sentence 'Washington County is the adjoins of what location? The answer is '." \\
        "The answer is Westmoreland County, so the [MASK] is Westmoreland County." \\
        \\
        \#\# \textbf{Demo 2}: "The question is to predict the head entity [MASK] from the given ([MASK], \textit{location adjoining\_relationship adjoins}, Rockland County) by completing the sentence 'Rockland County is the adjoins of what location? The answer is '."\\
        "The answer is Bergen County, so the [MASK] is Bergen County." \\
        \\
        \textcolor{darkblue}{\textbf{\#\# Task demonstrations with \textcolor{red}{Contextual Retrieval}}}: \\
        \\
        \#\# \textbf{Demo 1}: "\textcolor{red}{Washington County: county in Pennsylvania, U.S.} The question is to predict the head entity [MASK] from the given ([MASK], \textit{location adjoining\_relationship adjoins}, Washington County) by completing the sentence 'Washington County is the adjoins of what location? The answer is '." \\
        "The answer is Westmoreland County, so the [MASK] is Westmoreland County. \textcolor{red}{Westmoreland County: county in Pennsylvania, United States}" \\
        \\
        \#\# \textbf{Demo 2}: "The question is to predict the head entity [MASK] from the given ([MASK], \textit{location adjoining\_relationship adjoins}, Rockland County) by completing the sentence 'Rockland County is the adjoins of what location? The answer is '."\\
        "The answer is Bergen County, so the [MASK] is Bergen County. \textcolor{red}{Bergen County: county in New Jersey, United States}" \\
        \midrule
        \textcolor{darkblue}{\textbf{\#\# Candidate entities}}: [Cook County, Champaign, Bloomington, McHenry County, Evanston]\\
        \\
        \textcolor{darkblue}{\textbf{\#\# Candidate Answers with \textcolor{red}{Contextual Retrieval}}}: \\
        \textcolor{red}{Cook County: county in Illinois, United States}\\
        \textcolor{red}{Champaign County: county in Illinois, United States}\\
        \textcolor{red}{Bloomington: city and the county seat of McLean County, Illinois, United States}\\
        \textcolor{red}{McHenry County: county in Illinois, United States}\\
        \textcolor{red}{Evanston: suburban city in Cook County, Illinois, United States}\\
        \bottomrule
    \end{tabular}
    \caption{Prompt template of retrieval stage.}
    \label{tab:pt_retrieval}
\end{table}
\clearpage

\subsection{Prompt templates of Reasoning stage}\label{sec:appendix2}
Table~\ref{tab:pt_reasoning} shows the prompt templates of the \textbf{Reasoning} stage and give an example which is the same case as Table~\ref{tab:pt_retrieval}.
\begin{table}[h!]
    \centering
    \begin{tabular}{p{\textwidth}} % Change to full width
        \toprule
        \textcolor{darkblue}{\textbf{\#\# KG Triplet for completion}:}
        ([MASK], \textit{/location/adjoining\_relationship/adjoins}, Champaign) \\
        \\
        \textcolor{darkblue}{\textbf{\#\# Task for completion}:}
        "The question is to predict the head entity [MASK] from the given ([MASK], \textit{location adjoining\_relationship adjoins}, Champaign) by completing the sentence 'Champaign is the adjoins of what location? The answer is '."\\
        \midrule
        \#\# \textbf{\textcolor{darkblue}{Reasoning}:} \\
        \\
        The question is to predict the head entity [MASK] from the given ([MASK], \textit{location adjoining\_relationship adjoins}, Champaign) by completing the sentence 'Champaign is the adjoins of what location? The answer is '. Output all some possible answers based on your own knowledge, using the format '[answer1, answer2, ..., answerN]' and please start your response with 'The possible answers:'. Do not output anything except the possible answers. \\
        \\
        
        \#\# \textbf{\textcolor{red}{Context-aware} \textcolor{darkblue}{Reasoning}:} \\
        \\
        Here are some materials for you to refer to. \textcolor{red}{Champaign: Champaign is a city in Champaign County, Illinois, United States. The population was 88,302 at the 2020 census. It is the tenth-most populous municipality in Illinois and the fourth most populous city in the state outside the Chicago metropolitan area. It is a principal city of the Champaign–Urbana metropolitan area, which had 236,000 residents in 2020. Champaign shares the main campus of the University of Illinois with its twin city of Urbana, and is also home to Parkland College, which gives the city a large student population during the academic year. Due to the university and a number of technology startup companies, it is often referred to as a hub of the Illinois Silicon Prairie. Champaign houses offices for the Fortune 500 companies Abbott, Archer Daniels Midland (ADM), Caterpillar, John Deere, Dow Chemical Company, IBM, and State Farm. Champaign also serves as the headquarters for several companies, including Jimmy John's.} \\
        \\
        The question is to predict the head entity [MASK] from the given ([MASK], \textit{location adjoining\_relationship adjoins}, Champaign) by completing the sentence 'Champaign is the adjoins of what location? The answer is '. Output all the possible answers you can find in the materials using the format '[answer1, answer2, ..., answerN]' and please start your response with 'The possible answers:'. Do not output anything except the possible answers. If you cannot find any answer, please output some possible answers based on your own knowledge. \\
        \midrule
        \#\# \textcolor{darkblue}{\textbf{Context-aware Reasoning result by LLM}:} \\
        \\
        The possible answers: Urbana, Champaign County, Illinois Silicon Prairie, Parkland College.\\
        \bottomrule
    \end{tabular}
    \caption{Prompt Template of context-aware reasoning.}
    \label{tab:pt_reasoning}
\end{table}

\clearpage
\subsection{Prompt templates of Ranking stage}\label{sec:appendix3}
Table~\ref{tab:pt_ranking} shows the prompt templates of the \textbf{Retrieval} stage and give an example which is the same case as Table~\ref{tab:pt_retrieval} and~\ref{tab:pt_reasoning}. 

Noteworthily, this case also empirically shows the effectiveness of the \textbf{Reasoning} and \textbf{Re-ranking} processes. The ground truth answer 'Urbana' is not successfully retrieved by the KGC model, GIE. However, the LLM provides new candidates including the ground truth answer 'Urbana', by analyzing the context of the known entity 'Champaign' in the incomplete triple during the \textbf{Reasoning} process. And the LLM succeed in re-ordering the whole candidate list based on the contexts of candidates and giving the correct answer during the \textbf{Re-ranking} process. 
\begin{table}[h!]
    \centering
    \begin{tabular}{p{\textwidth}} % Change to full width
        \toprule
        \textcolor{darkblue}{\textbf{\#\# KG Triplet for completion}:}
        ([MASK], \textit{/location/adjoining\_relationship/adjoins}, Champaign) \\
        \\
        \textcolor{darkblue}{\textbf{\#\# Task for completion}:}
        "The question is to predict the head entity [MASK] from the given ([MASK], \textit{location adjoining\_relationship adjoins}, Champaign) by completing the sentence 'Champaign is the adjoins of what location? The answer is '."\\
        \midrule
        \#\# \textbf{\textcolor{darkblue}{Re-Ranking}:} \\
        \\
        The question is to predict the head entity [MASK] from the given ([MASK], \textit{location adjoining\_relationship adjoins}, Champaign) by completing the sentence 'Champaign is the adjoins of what location? The answer is '. The list of candidate answers is [Cook County, Champaign County, Bloomington, Evanston, Urbana]. \\
        Sort the list to let the candidate answers which are more possible to be the true answer to the question prior. Output the sorted order of candidate answers using the format '[most possible answer, second possible answer, ..., least possible answer]' and please start your response with 'The final order:'. \\
        \\
        \#\# \textbf{\textcolor{red}{Context-aware} \textcolor{darkblue}{Re-Ranking}:} \\
        \\
        \textcolor{red}{Champaign: city in Champaign County, Illinois, United States}\\
        The question is to predict the head entity [MASK] from the given ([MASK], \textit{location adjoining\_relationship adjoins}, Champaign) by completing the sentence 'Champaign is the adjoins of what location? The answer is '. The list of candidate answers is [Cook County, Champaign County, Bloomington, Evanston, Urbana]. \\
        \textcolor{red}{Cook County: county in Illinois, United States}\\
        \textcolor{red}{Champaign County: county in Illinois, United States}\\
        \textcolor{red}{Bloomington: city and the county seat of McLean County, Illinois, United States}\\
        \textcolor{red}{McHenry County: county in Illinois, United States}\\
        \textcolor{red}{Evanston: suburban city in Cook County, Illinois, United States}\\
        \textcolor{red}{Urbana: town in and county seat of Champaign County, Illinois, United States}\\
        Sort the list to let the candidate answers which are more possible to be the true answer to the question prior. Output the sorted order of candidate answers using the format '[most possible answer, second possible answer, ..., least possible answer]' and please start your response with 'The final order:'. \\
        \midrule
        \#\# \textcolor{darkblue}{\textbf{Re-Ranking Result generated by LLM}:} \\
        \\
        The final order: [Urbana, Champaign County, Cook County, Bloomington, McHenry County Evanston] \\
        \\
        \#\# \textcolor{darkblue}{\textbf{Evaluation}:} The ground truth 'Urbana' hits at 1\\
        \bottomrule
    \end{tabular}
    \caption{Prompt Template of context-aware ranking.}
    \label{tab:pt_ranking}
\end{table}

\end{document}

%% file: intro.tex
\section{Introduction}
Knowledge Graphs~(KGs) are structured knowledge bases~(KBs) that organize factual knowledge as triples in the form of \textit{(head entity, relation, tail entity)}. These triples interweave into a graph-like structure, where each node represents an entity and each edge represents a relationship. 
% In the process of constructing a KG, it is necessary to first design the ontology of the KG. The ontology defines the categories of entities, properties, and relationships, as well as their hierarchical structure, thereby ensuring that the data instances in the KG have a consistent structure that is easy for computers to understand and utilize. 
This structured representation enables machines to easily understand and reason about knowledge, thereby supporting various intelligent applications such as question answering~\cite{ToG}, semantic analysis~\cite{wang-shu-2023-explainable}, recommendation systems~\cite{Recommendation}, and more. 

\begin{figure*}[t!]
    \centering
    \includegraphics[width=\textwidth]{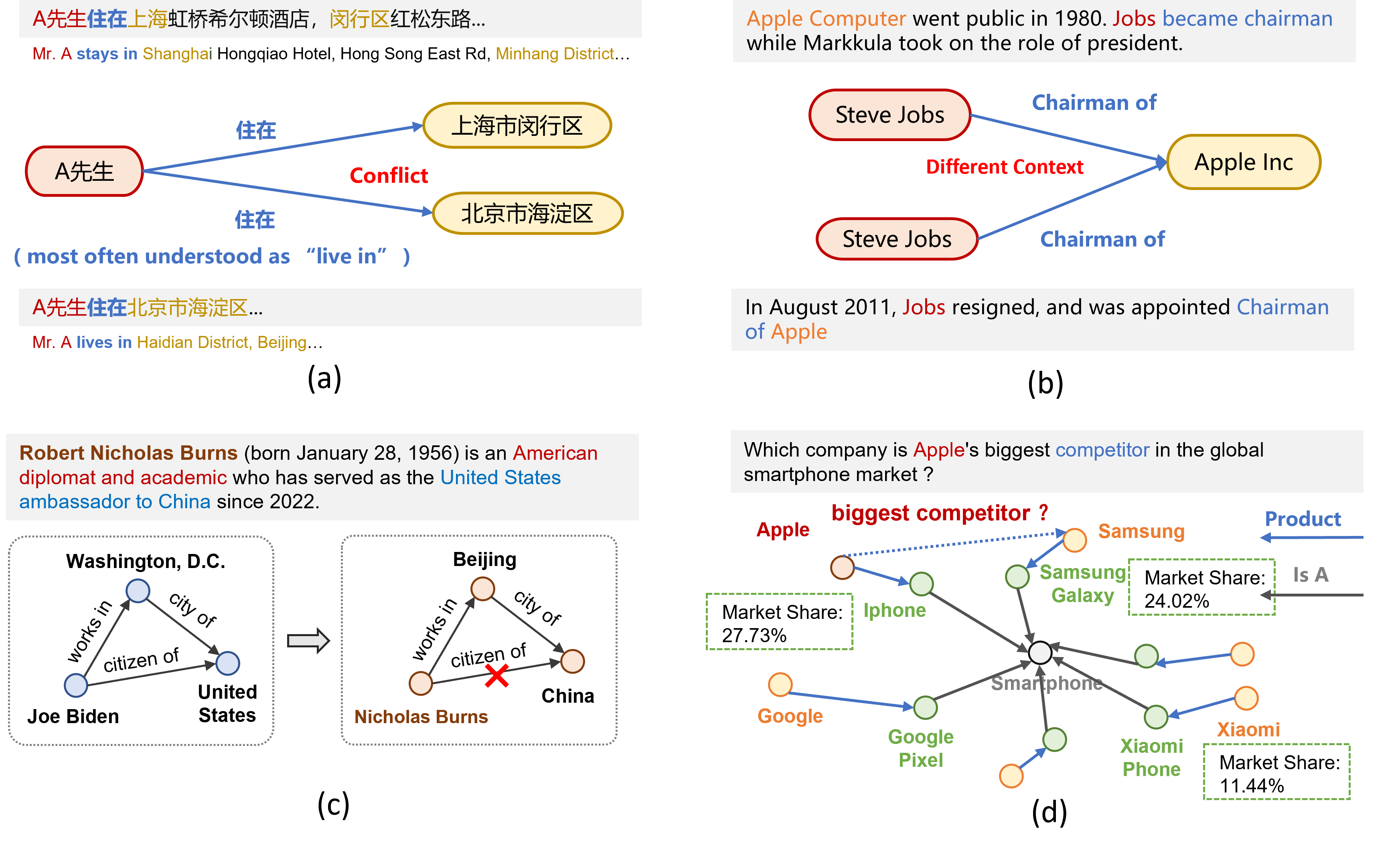}
    \vspace{-0.9cm}
    \caption{Examples of limitations of triple-based KGs. (a) gives an example that the loss of contextual information during KG construction processes may result in the extraction of contradictory triples; (b) gives an example that triple-based representation struggle to represent two facts that involve the same entities and relations but occur in different contexts; (c) gives an example that triple-based KG reasoning methods often learn rule patterns that frequently occur in KGs, but they tend to ignore contexts that may affect the validity of these rules; (d) gives an example that triple-based KG reasoning methods face difficulties in answering questions that involve relational knowledge or contextual information beyond the scope of the triples in KGs.}
    \label{fig:tripleKG}
\end{figure*}

While this triple-based structure offers clear semantics and precision through the use of schemas and ontologies, it loses the contextual information of knowledge and falls short in capturing the complexity and richness of real-world knowledge~\cite{textKG}. Since we cannot clearly model the knowledge in a domain only with entities and relations, many recent KGs~\cite{yago4,wikidata} are designed to be semi-structured: they leverage the clear semantics of structured data provided by the rigidity of schemas (i.e., ontologies) while also embracing the flexibility of unstructured data. Such KGs integrate multi-modal knowledge, including entity description, images, timestamps and other metadata, all of which can be regarded as the contexts of triple knowledge. In this paper, we refer to this type of KGs as contextual graphs (CGs). By incorporating these semantic contexts, CGs provide a more comprehensive and nuanced representation of knowledge, extending beyond the traditional triple-based approach. This enables KGs to possess more advanced capabilities in knowledge representation and reasoning.

Moreover, large language models (LLMs), pre-trained on vast text corpora, have exhibited strong semantic understanding capability~\cite{GPT3}. And the use of LLMs for KG reasoning has become a research hotspot~\cite{KICGPT,GenTKG,ToG}. However, KGs may contain numerous entities and relations, but not all entities and relationships are fully annotated and connected, leading to data sparsity. This sparsity results in a lack of sufficient contextual information for the LLM during inference. On the other hand, LLMs are better at handling unstructured data rather than structured triples. Considering that CGs can provide unstructured contextual information for LLM reasoning, the synergy between CGs and LLMs holds significant potential for advancing the field of knowledge reasoning.

In this paper, we will first give a brief discussion on the limitations of the triple-based KGs and give the specific definition of CGs. To validate the effectiveness of contexts on enhancing knowledge representation and reasoning, we propose a novel context graph reasoning paradigm, named \textbf{CGR}$^{\mathbf{3}}$, which leverage the strong reasoning power of LLMs to firstly \textbf{retrieve} candidate entities and related contexts from KG, and \textbf{rank} the candidate entities based on retrieved context, and then \textbf{reason} whether sufficient information is retrieved to answer the question. Experimental results demonstrate that our proposed paradigm CGR$^{3}$ enhances the performance of existing models on the tasks of KG completion (KGC) and KG question answering (KGQA), which are two of fundamental reasoning tasks over KGs. 

Overall, this paper have two major contributions:
\begin{itemize}
    \item Point out the limitations of the current triple-based KGs, and introduce the concept of Context Graph, which has more advanced capabilities in knowledge representation and reasoning.
    \item Propose a context-enhanced KG reasoning paradigm, CGR$^3$, which leverages the LLM to perform CG reasoning based on related contexts. Experimental results on KGC and KGQA support our intuition that the integration of contextual data can contribute to effective KG reasoning.
\end{itemize}

%% file: definition.tex
\section{Context Graph}
In this section, we first discuss on the limitations of triple-based KGs, caused by the absence of contextual information. Moreover, we point out the effects of contextual information on knowledge representation and reasoning, then categorize and interpreter different types of contexts in the KGs. Finally, we formally define CGs as well as two knowledge reasoning tasks over CGs.

\subsection{Limitations of Triple-based KGs}
A Triple-based Knowledge Graph (denoted as $\mathcal{KG} = \{\mathcal{E}, \mathcal{R}, \mathcal{T}\}$) can be represented as a set of triples in the form of $(h, r, t)\in \mathcal{T}$, where $h, t \in \mathcal{E}$, $r \in \mathcal{R}$. The notations $h$ and $t$ denote the head and the tail entity of a triple. $\mathcal{E}, \mathcal{R},\mathcal{T}$ are the set of entities, relations, and triples, respectively. Typical triple-based KGs include Freebase~\cite{Freebase}, WordNet~\cite{WordNet} and DBPedia~\cite{DBPedia}. 
In these triple-based KGs, the triple representation excludes crucial contextual information, often resulting in inaccurate knowledge storage, incomplete representation, and ineffective reasoning. These issues are the primary constraints on the practical application of most current KGs.

\begin{CJK}{UTF8}{gbsn}
To be specific, the same relationship may have different meanings in different contexts, thus the triple representation could lead to incorrect knowledge storage. For instance, consider the two sentences: '\textit{A先生住在上海虹桥希尔顿酒店，闵行区红松东路}' and '\textit{A先生住在北京市海淀区}' as shown in Figure~\ref{fig:tripleKG}(a). They may be represented as two triples: \textit{(A先生, 住在, 上海市闵行区)} and \textit{(A先生, 住在, 北京市海淀区)}, respectively, in a KG. However, these representations are semantically conflicting since a person cannot live in two places simultaneously. This mistake is likely to occur because the predicate '\textit{住}' in the first sentence implies '\textit{stay in}', whereas in the latter one, it denotes '\textit{live in}'. The triple extraction process filters out the sentence context, leading to information conflicts. 
\end{CJK}

Moreover, each data instance in a KG strictly adheres to its ontology structure. The ontology structure defines the categories of entities, relations, and attributes, as well as their hierarchical relationships. During the construction of a KG, knowledge outside the pre-defined categories is filtered out, including a large amount of contextual information, leading to incomplete knowledge representation. For example, the contexts of Steve Jobs serving as the chairman of Apple Inc. twice are very different as shown in Figure~\ref{fig:tripleKG}(b). However, based on triple representation, both events would be represented as (\textit{Steve Jobs, chairman of, Apple Inc.}), which results in downstream tasks not obtaining sufficient information when utilizing related knowledge.

Triple-based knowledge representation also limits the effectiveness of existing KG reasoning methods, which mainly focus on learning explicit or implicit rules through rule mining or embedding models. For example, from triple (\textit{X, works in, Y}) and (\textit{Y, city of, Z}), it is very likely for KG reasoning models to deduce that (\textit{X, citizen of, Z}) since such rule pattern appears frequently in the training data as shown in Figure~\ref{fig:tripleKG}(c). However, these probability based rules may not hold in all contexts, leading to conclusions that do not align with the facts. Besides, triple-based KGs only contain relational knowledge limited by predefined relation set $\mathcal{R}$. The triple-based reasoning process have difficulties in answering questions involving relations out of $\mathcal{R}$ without additional contextual information or external data sources. 

% % 由于KG是不完整的，相关联的entity并不一定会被直接或间接地通过certain type of relation所连结。因此，KG可能并不能对query查询提供确实可靠的证据。
% due to the incompleteness nature of KGs, the related entities might not be directly or indirectly connected by the certain type of relations. Consequently, KGs may not be able to provide reliable support for all queries. For instance, consider question ``\textit{Who was the secretary of state when Andrew Jackson was president?}''. It will be difficult for triple-based reasoning models to predict the ground truth \textit{James A. Hamilton} if the entities \textit{James A. Hamilton} and  \textit{Andrew Jackson} are not conntected directly.
% % Consequently, existing methods may fail to capture and utilize the common salient properties of different entities. 
% % 这个例子是从数据集里抽的真实例子，知识图谱里没有证据表明James A. Hamilton为Andrew Jackson工作过。即便他们互相间接相连，但我们还是无法得出答案。
% For instance, consider question ``Who was the secretary of state when andrew jackson was president?'' \textit{limiting their ability to perform analogical reasoning}. This may compromise the reliablity of downstream applications. 
% %perform analogical or similarity-based reasoning, which restricts their capability for leapfrogging thinking. These limitations highlight the need for future research to overcome the structural and completeness constraints of knowledge graphs and to better integrate and utilize semantic information for deeper reasoning.

\subsection{The Effects of Contextual Information}
To address the limitations of triple-based KGs, a promising approach is to attach contextual data to factual triples. For instance, several KGs, such as YAGO and the Yahoo Knowledge Graph, include meta-information with their facts, such as the time of validity, the geographic location of a fact, and provenance information. By integrating such data, CGs can offer a more comprehensive and accurate representation of knowledge, thereby enabling more effective reasoning.

\paragraph{Knowledge Representation:} Contextual data provide additional layers of information that enhance the representation and understanding of facts. For example, contextual data can differentiate facts that have the same relations and entities but occur in different backgrounds, such as recurring events in history. This differentiation allows for a more nuanced and detailed understanding of the information, capturing the various dimensions in which similar facts can differ based on time, location, and other contextual elements.

\paragraph{Knowledge Reasoning:} During the process of knowledge reasoning, contextual information within CGs can be leveraged to associate entities that are not directly connected by identifying similar contexts. This capability is particularly useful for making connections and drawing inferences that go beyond the predefined relation set of a triple-based KG. Moreover, contextual information provides additional knowledge, allowing for larger knowledge coverage and greater flexibility compared to triple-based KGs. Specifically, contextual information can be used to answer complex reasoning questions, such as those involving qualifiers or specific conditions that are often hidden within contextual data. For instance, answering a question about "\textit{which company is Apple's biggest competitor in the global smartphone market}" would require integrating quantitative data, temporal information, and detailed market dynamics analysis with basic entity and relation information in KGs, as shown in Figure~\ref{fig:tripleKG}(d). CGs thus enable the handling of such intricate queries by providing a richer and more detailed knowledge base.

\subsection{Categories of Contextual Data}
\begin{figure}
    \centering
    \includegraphics[width=.48\textwidth]{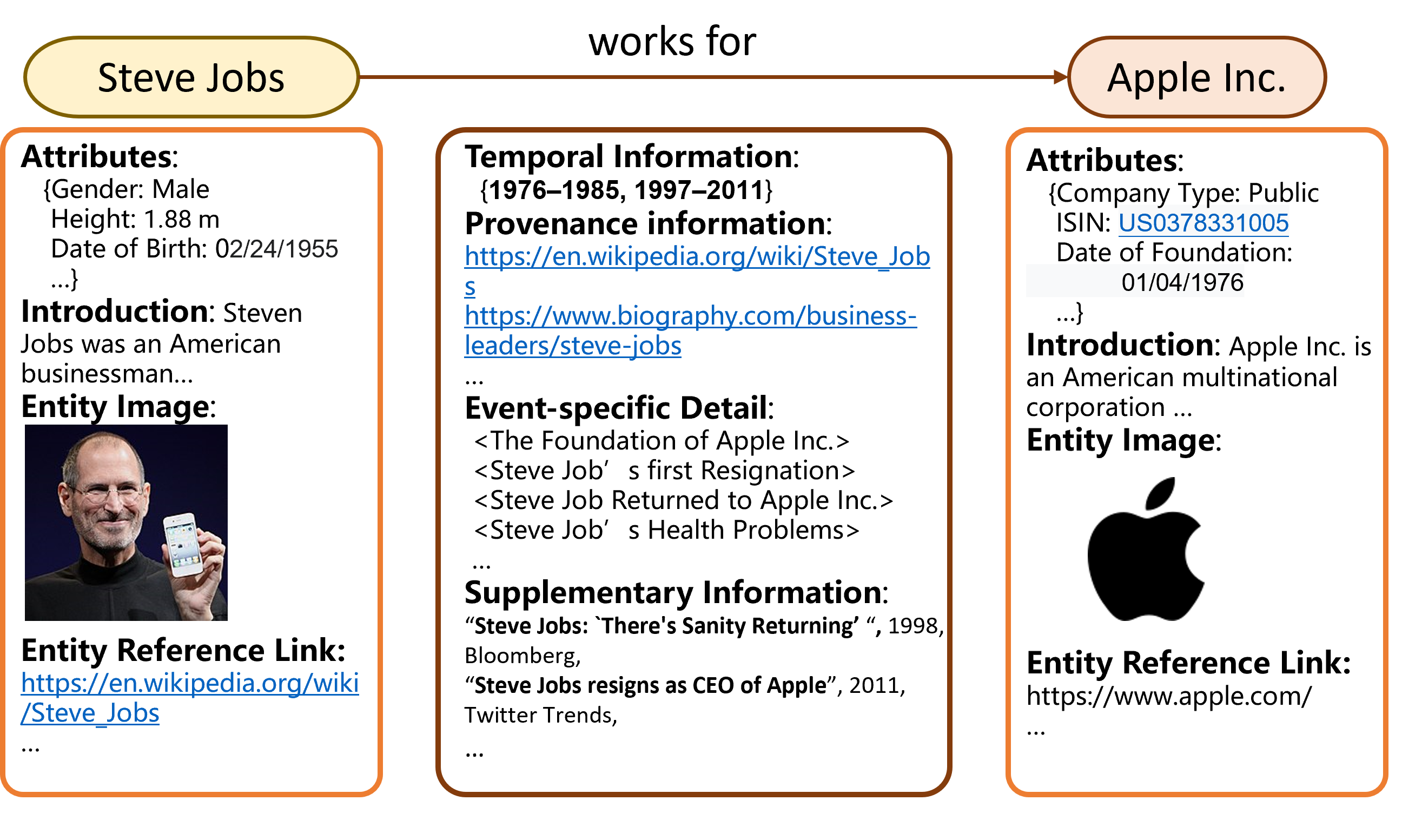}
    \caption{An example of factual triples with entity and relation contexts}
    \label{fig:contextKG}
\end{figure}

% \usepackage{multirow}

\begin{table*}[t!]
\centering
% \resizebox{\textwidth}{!}{
\begin{tabularx}{\textwidth}{p{2.65cm}|p{3.35cm}|p{4.5cm}|p{3.8cm}}
\hline
          Category        & Context Type & Description   & Instance \\ \hline
\multirow{20}{*}{Entity Context} &Entity Attribute &Specific properties or characteristics of the entity& Person: height, gender Product: price, color\\ \cline{2-4} 
                  &Entity Type  & Classifications or types to which the entity belongs, providing context within a larger framework or ontology.  &Person: actor, artist, scientist, athlete, musician  \quad \quad \quad  Place: landmark, city, country, state  \\ \cline{2-4} 
                  &Entity Description  &Textual descriptions that provide a comprehensive overview of the entity.  &Person: A detailed biography or background  \\ \cline{2-4} 
                  &Entity Alias  &Alternative names or identifiers for the entity.  & Istanbul, alias: Constantinople.  \\ \cline{2-4} 
                  &Entity Reference Link&Links to external resources or webpages that provide additional information about the entity.  &Wikipedia pages, official websites, social media profiles, etc.  \\ \cline{2-4} 
                  &Entity Image  &Visual representations or photographs of the entity.  & Person: photographs or portraits \\ \cline{2-4} 
                  &Entity Speech  &Audio recordings or sounds associated with the entity.  &Music audio, audio introductions, etc. \\ \cline{2-4} 
                  &Entity Video  &Video clips or recordings that feature the entity.  & Video interviews, a TED talk, etc.  \\ \hline
\multirow{20}{*}{Relation Context} &Temporal Information  &The time period during which a relationship is valid or relevant.  & (Barack Obama, president of, USA, time: 2009-2017) \\ \cline{2-4} 
                  &Geographic Location  &The physical location associated with a relationship or an event involving entities.  &(France national football team, win, 2018 FIFA World Cup, location: Russia) \\ \cline{2-4} 
                  &Quantitative Data  &Specific numerical or quantitative information directly related to the relationship.  &(Berkshire Hathaway, shareholder of, Apple Inc,  Quantity: 790 million shares)  \\ \cline{2-4} 
                  &Provenance information &References to the origin or source of the relationship data.&Documents, news, articles, images, datasets, etc.  \\ \cline{2-4} 
                  &Confidence Level  & Indicators of the reliability or confidence in the relationship data. &The accuracy of the relation extraction model  \\ \cline{2-4} 
                  &Event-specific Detail &Information about specific events that define or influence the relationship between entities.  &(Argentina national football team, win, France national football team, event: 2022 FIFA World Cup)  \\ \cline{2-4} 
                  &Supplementary Information  &Information that provides background or additional context to the relationship, explaining its significance or implications.  &News topics, comments, read counts, share counts, like counts, etc.\\ \hline
\end{tabularx}
\caption{Examples of different types of entity and relation contexts.}~\label{tb:context type}
\end{table*}

As shown in Figure~\ref{fig:contextKG}, contextual data  can be roughly classified into two categories, i.e., entity contexts and triple contexts.

Entity contexts refer to information that provides a deeper understanding of an individual entity within the KG. This type of context helps in defining the attributes, characteristics and backgrounds of the entity. Entity contexts include entity attributes, entity types, entity descriptions, entity aliases, entity reference links, entity images, entity speeches, entity videos, etc. 

Relation contexts refer to specific pieces of information that describe the relations between entities. They provides concrete data points and factual statements that contribute to the KG's informational content. Relation contexts include temporal information, geographic locations, quantitative data, provenance information, confidence levels, event-specific details, and other supplementary information. By incorporating these relation contexts, KGs can offer a richer, more detailed representation of the relationships between entities, enhancing their overall accuracy and utility for reasoning and analysis.

Table~\ref{tb:context type} demonstrates some examples of different types of entity contexts and relation contexts.

\subsection{Problem Specification}
A Context Graph (denoted as  $\mathcal{CG} = \{\mathcal{E}, \mathcal{R}, \mathcal{Q}, \mathcal{EC}, \mathcal{RC}\}$) can be represented as a set of factual quadruples in the form of $(h, r, t, rc)\in \mathcal{Q}$, where $h, t \in \mathcal{E}$, $r \in \mathcal{R}$ and $rc \in \mathcal{RC}$. The notations $h$ and $t$ denote the head and the tail entity of a factual quadruple, $r$ denotes the relations between $h$ and $t$, and $rc$ denotes . $\mathcal{EC}, \mathcal{RC}$ are the set of entity contexts and relation contexts. Each entity $e \in \mathcal{E}$ and its entity context $ec \in \mathcal{EC}$ form a complete entity representation $(e,ec)$.

To validate whether contextual information can be used to enhance the ability of KG reasoning models, in this paper, we consider two KG reasoning tasks for verification, i.e., KG completion (KGC) and KG question answering (KGQA). 
\paragraph{Knowledge Graph Completion}
Given a query $(h, r, ?)$ or $(?, r, t)$, KGC aims to predict the missing tail or head entity (denoted as ``?'') that will make the quadruple plausible when the relation context is unknown. Based on the convention of ranking-based evaluation metrics, the aim of a KGC model is to learn a scoring function $f(h,r,t)$ to measure the plausibility of all entities in $\mathcal{E}$ as the missing ones in the query and then rank them in descending order. For performing KGC over a contextual KG, the scoring function $f(h,r,t)$ can be reformulated as $f(h,r,t,hc,rc,tc)$, where $hc\in\mathcal{EC}$, $rc\in\mathcal{RC}$, $tc\in\mathcal{EC}$ denote the contexts of head entity, tail entity and the relation between them, respectively. 
\paragraph{Knowledge Graph Question Answering} Given a natural question $nq$ and its topic entity $e_{topic}\in\mathcal{E}$, KGQA aims to retrieve related knowledge by generating structured queries or sampling subgraphs from $\mathcal{KG}$ and predict the answer $a$ based on retrieved knowledge, i.e, $a = f(nq,e_{topic},\mathcal{KG})$. For performing QA over a contextual KG, the prediction function can be reformulated as $f(nq,e_{topic},\mathcal{CG})$. 